\documentclass[journal]{IEEEtran}

\usepackage{graphicx}
\usepackage{caption}
\usepackage{subcaption}
\usepackage[ruled,vlined]{algorithm2e}
\usepackage{algorithmic}
\usepackage{amssymb}
\usepackage{amsmath}
\usepackage{empheq}

\begin{document}
	
\title{A Safety-Critical Decision Making and Control Framework Combining Machine Learning and Rule-based Algorithms}

\author{
Andrei~Aksjonov~and~Ville~Kyrki\

Intelligent Robotics Group, Department of Electrical Engineering and Automation,\

Aalto University, 02150 Espoo, Finland\

(e-mails: andrei.aksjonov@aalto.fi; ville.kyrki@aalto.fi)
}

\markboth{}{}
\maketitle

\begin{abstract}
While artificial-intelligence-based methods suffer from lack of transparency, rule-based methods dominate in safety-critical systems. Yet, the latter cannot compete with the first ones in robustness to multiple requirements, for instance, simultaneously addressing safety, comfort, and efficiency. Hence, to benefit from both methods they must be joined in a single system. This paper proposes a decision making and control framework, which profits from advantages of both the rule- and machine-learning-based techniques while compensating for their disadvantages. The proposed method embodies two controllers operating in parallel, called Safety and Learned. A rule-based switching logic selects one of the actions transmitted from both controllers. The Safety controller is prioritized every time, when the Learned one does not meet the safety constraint, and also directly participates in the safe Learned controller training. Decision making and control in autonomous driving is chosen as the system case study, where an autonomous vehicle learns a multi-task policy to safely cross an unprotected intersection. Multiple requirements (i.e., safety, efficiency, and comfort) are set for vehicle operation. A numerical simulation is performed for the proposed framework validation, where its ability to satisfy the requirements and robustness to changing environment is successfully demonstrated.\
\end{abstract}


\section{Introduction}

\IEEEPARstart{A}{ human} being is prone to fatigue, drowsiness, distraction and many other human-factor-based characteristics that often lead to dramatic accidents. Artificial Intelligence, however, has a great potential to replace people in safety-critical decision making and control systems, where human-factor-induced mistakes are no longer acceptable. Artificial Intelligence in many ways mimics human learning with knowledge acquisition and skill refinement~\cite{Michalski83}.\
	
	Machine Learning (ML)-based systems, however, suffer from lack of decision process transparency (i.e., they commonly serve as "black-box")~\cite{Yoon94}. Hence, these methods often experience challenges in safety-critical applications, such as Autonomous Vehicle (AV), because it is of a great importance to track every part of an autonomous decision making and control to prevent accidents induced by design or optimization errors. Nevertheless, ML is robust to complexity and capable of efficient policy generalization and scalability~\cite{Likmeta20}.	
				
\begin{figure}[!t]
	\centering
	\includegraphics[width=3.4in]{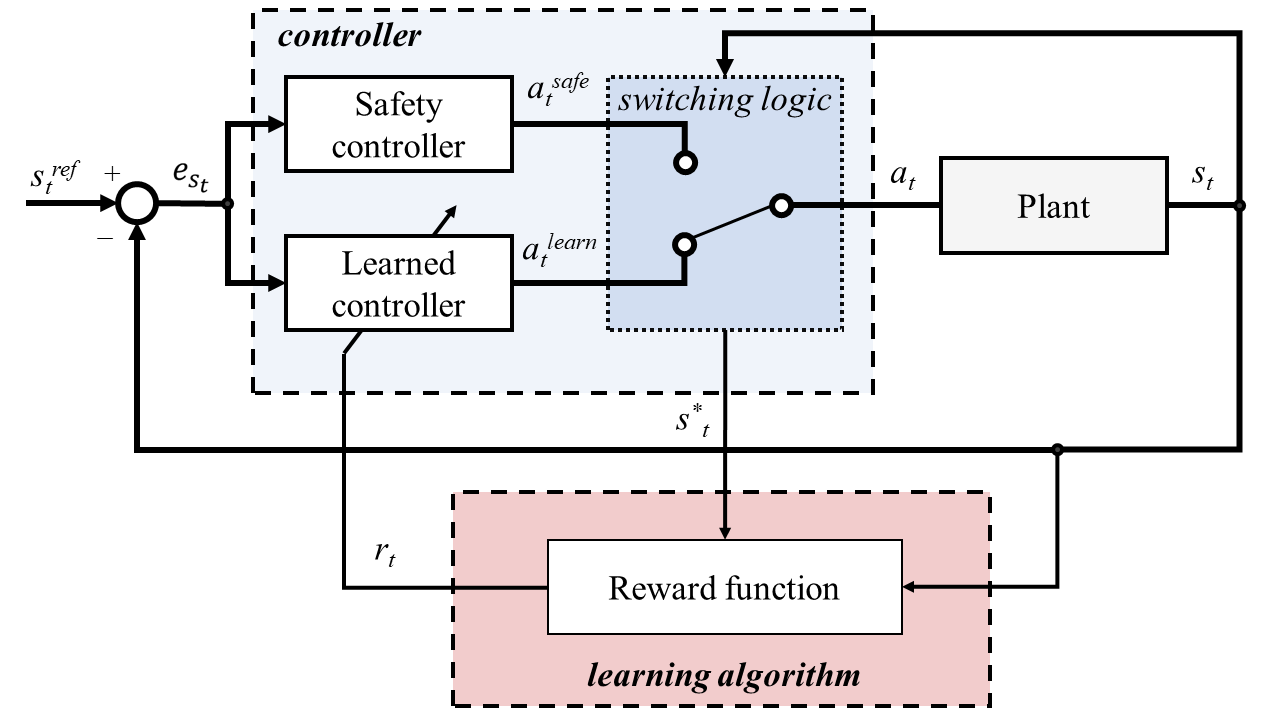}
	\caption{Decision making and control system block scheme: $ a $ - action; $ a^{safe} $ - action from Safety controller; $ a^{learn} $ - action from Learned controller; $ r $ - reward; $ s $ - state of the plant; $ s^{*} $ - state of the \textit{switching logic}; $ s^{ref} $ - reference state; $ e_{s} $ - error of the state; $ t $ - time step.}
	\label{fig_1}
\end{figure}
	
On the contrary, Rule-Based (RB) methods dominate in safety-critical systems. This is primarily thanks to "readability" characteristic: they are qualified for transparent and understandable decision making and control definition. Since human knowledge is implicit~\cite{Polanyi}, RB methods in comparison to ML, fail in fulfilling multiple independent criteria (e.g., safety, efficiency, comfort) and complex scalability of a plant or in accounting for uncertainties. Furthermore, design of universal handcrafted rules is extremely complicated and prone to hidden errors~\cite{Likmeta20}.\
	 
	Combining RL and ML methods aims at utilizing strong parts of each technique in sophisticated plant, where both transparency and scalability are required. Incorporation of RB safety constraints in ML unlocks learning control of safety-crucial systems~\cite{Likmeta20},~\cite{Hewing20}. Yet, a comprehensive decision making and control architecture that reaches multiple plant performance requirements, which ensures not only safe utilization of an ML model but also allows for safe training of the agent, is desired today~\cite{Maravall09}.\
	
	The main contributions of this paper are three-fold:
	
\begin{itemize}
	\item A safety-critical decision making and control framework combining RB and ML techniques (Fig.~\ref{fig_1}) is proposed, which gains from the advantages of both (i.e., transparency and scalability), where safety requirements of the system are fulfilled not only during the model utilization but also while training.
	\item The ML-based controller is designed to concurrently address multiple requirements (i.e., safety, comfort, and efficiency), while active in parallel RB Safety controller ensures safety at any cost and also facilitates continuous learning and ML policy enhancement in the long term.
	\item The proposed framework is adapted to and validated on AV performing a safety-critical unprotected left-turn maneuver on cross-intersection learning a multi-task policy.
\end{itemize}
	
	The rest of the paper is structured in the following way. Section II analyses related works and lists the distinctive elements of the proposed framework with respect to them. Section III describes the developed system. A case study on AV application is explained in Section IV. The training results of the Learned controller are shown in Section V. Section VI is dedicated to framework's experimental verification. The paper is concluded in Section VII.\\
	
\section{Related Works}
	
	The related works in the area can be categorized to hybrid methods (the ML approaches are accompanied by RB or computational intelligence  algorithms), game theoretic (GT), (deep) Reinforcement Learning (RL) (relies on expert's knowledge expressed in reward functions), and safe RL.\
	
\subsubsection{Hybrid methods}
	
	Computational intelligence community has addressed in depth the challenge of combining RB and ML ~\cite{Maravall09}. A parameter setting mechanism for a RB evolutionary ML system that is capable of finding the adequate parameter value for a wide variety of synthetic classification problems with binary attributes is presented in~\cite{Franca20}. He \textit{et al.}~\cite{He20} promoted an on-line RB classifier learning framework on dynamically added unlabeled data. Lexicographic Optimization-based model predictive controller ethical decision making system was presented in~\cite{Wang20}. The mentioned solutions are powerful in algorithm transparency and continuous optimization, yet they are not solving problem of scalability, what is essential in modern robotic systems, especially, when complexity of the system constantly increases.\
	
	Fuzzy Logic (FL) is a powerful tool in modeling expert's knowledge. Eventually, FL was efficiently incorporated with other ML methods~\cite{Bhattacharyya15}. A hybrid learning algorithm includes training data set with type-2 FL Artificial Neural Network (ANN)~\cite{Yeh11}. In \cite{Khanesar12}, the parameters of type-2 FL system were trained using extended Kalman filter. The RL proves to be one of the most powerful tools for optimizing a controller~\cite{Ibrahim20}. Another FL Q-learning method was tested on a complex cooperative multi-agent system~\cite{Kofinas18}. Nevertheless, satisfying multiple requirements of the process is still challenging for these systems. Also, they are mainly used in controller tuning rather than decision making system development.\
	
\subsubsection{Game theory}
	
	Coming from economics discipline, the cooperative and non-cooperative GT is widely used as a decision making system in complex multi-agent environments for simultaneously addressing safety, rapidity, and comfort indicators~\cite{Hang20}. A GT-based controllers for safety-critical AV's trajectory following were developed in~\cite{An21}. A GT model based on the concept of Nash equilibrium was also widely used in decision making~\cite{Wang21}. The biggest bottleneck in applying GT in practice is an uncertainties induced by other agents behavior prediction. This becomes especially challenging in safety-critical systems. Regarding for example AV, vehicle-to-vehicle communication, what noticeably improves accuracy of the prediction, is still futuristic, not to mention that the first AVs will operated in mixed traffic.
	
\subsubsection{(Deep) reinforcement learning}
	
	An ANN trained with RL is widely used as a decision making and control model. A popular approach is to shape the ANN parameters directly from a human expert's knowledge~\cite{Jagodnik17}. More advanced methods combine RL with evolutionary computation~\cite{Baldominos19} or model predictive control~\cite{Koller19}. A RL-based online evolving framework is proposed to detect and revise a controller's imperfect decision-making in advance in~\cite{Han20}. Deep RL is characterized with performance enhancement on complex tasks~\cite{Kiran2021}. Hoe \textit{et al.}~\cite{Hoel20} developed a general framework for tactical decision making, which combines the concepts of planning and learning. In~\cite{Chen19}, a model-free deep RL in challenging urban AV driving scenarios was developed. A comprehensive survey on deep RL on AV can be found in~\cite{Kiran2021}. Although demonstrated results are promising, all mentioned approaches are not applicable in safety-critical systems. Namely, they allowed the agent to make mistakes while training in order to learn, how not to make mistakes in model exploitation. For AV, for example, it means that it will need to collide multiple iterations to learn that the collision must be avoided. This is not practically feasible, because the cost of the training will be very high, let alone injuries and deaths it will lead to during the training, unless a safety driver is added for ensuring RL's safe action~\cite{Kendall19}, what in fact only increases the cost of the system.\
	
\subsubsection{Safe (deep) reinforcement learning}
	
	Ultimately, research evolved in a novel direction on exploring safe (deep) RL techniques. The scholars in~\cite{Isele19} investigated how to use prediction for safe learning. Addressing the challenges of existing RB and RL approaches, a modular decision making algorithm was proposed in~\cite{Bouton19}. A safety system consisting of two models (i.e., RB heuristic module, and data-driven ML-based one) was described in~\cite{Baheri20}. A safe RL algorithm, The Parallel Constrained Policy Optimization, that ensures the policy is safe in the learning process and improves the convergence speed was described in~\cite{Wen20}. A model-free offline RL-based approach with incorporated safety verification was developed in~\cite{Mirchevska18}.	Deep RL must be also concerned about safety. A combination of deep RL with expert's demonstration applied to AV's motion control was presented~\cite{Liu2021}. In~\cite{Ferdowsi18}, an adversarial deep RL for robust control considering safety and security was proposed. Authors in~\cite{Xiong16} developed a safety-based controller using an artificial potential field for deep RL. In safe RL, safety supervisors, e.g., human driver models, are utilized for safe exploration and exploitation assurance. These models intend to predict human behavior, and hence, minimize the level of uncertainty. However, these works surprisingly report a number of collisions after the training. This means that the methods are not able to ensure safety even after a model converges, not to mention during the training. Therefore, they are only approximately safe~\cite{Wen20}.\
	
	Furthermore, most of the decision making and control systems are expected to meet multiple independent requirements. For instance, incorporation of a human decision making model in RL for safe and efficient plant control was described in~\cite{Chen20}, which is built on data gathered from human performance. In~\cite{Shalev-Shwartz16}, the RL learned comfortable and safe control in multi-agent scenario. In~\cite{Zhu20}, multiple features, such as safety, efficiency, and comfort, were tackled in velocity control developed with deep RL. Xu \textit{et al.} \cite{Xu20} developed a RL approach with value function approximation and feature learning for autonomous decision making of intelligent vehicles simultaneously considering safety, smoothness, and speediness.\
	
\subsubsection{Distinctive features of the proposed framework}
	
	In this paper, a safety-critical decision making and control framework (Fig.~\ref{fig_1}) is proposed, which with respect to related works has the following features at once:
\begin{itemize}
	\item with transparently defined RB model guarantees \textbf{safety} not only after learned model convergence, but likewise during the training procedure;
	\item unlocks a \textbf{multi-task policy learning} or an existing one improvement over time, hence, reaction on uncertainty of the system;
	\item validated in AV case study, supports \textbf{scalability}: not only safety of the AV and environment, but provides efficiency and comfort at the same time.\\
\end{itemize}
	
\section{Decision-Making and Control System}

	The proposed decision making and control system is presented as a block scheme in Fig.~\ref{fig_1}. It is composed of two main parts: \textit{learning algorithm} and \textit{controller} with a decision making \textit{switching logic}. After successful learned controller convergence, the \textit{learning algorithm} may be turned off from the loop. In fact, it may also remain for continuous learning purpose.\
	
	\textit{Remark 1:} In this paper, due to the nature of the case study, continuous learning is not explored. Hence, the \textit{learning algorithm} is removed from the loop after the training procedure.\
	
\subsection{Controller}
	
	The \textit{controller} contains two controllers: the Safety and the Learned ones. The first controller tends to guarantee overall system safety under any situation. Therefore, to ensure its robustness and functionality it is designed as a rule- or model-based function. The Safety controller prioritizes safety, thus, it may easily sacrifice all together efficiency and comfort.\
	
	The Learned controller is a data-based model, for instance, ANN. It requires either preliminary collected data or a random data sample approach, e.g., RL. The latter is in particular highlighted in Fig.~\ref{fig_1}, which includes Reward function that modifies the shape of the Learned controller after each time step $ t $. The Safety controller also participates in Learned controller training via additional state of the system.\
		
	The \textit{controller} serves as a closed-loop control scheme. It receives error $ e_{s} $ calculated as a difference between reference state $ s^{ref} $ and actual state $ s $ at time $ t $. The $ s $ consists of one or multiple variables that are directly measured by appropriate sensors or estimated from available signals. The \textit{controller} releases control signal or action $ a $ straight to the Plant.\
	
	The Safety and the Learned controllers simultaneously function in parallel, they output safe $ a^{safe} $ and learned $ a^{learn} $ actions, respectively. However, only one signal is fed to the Plant at each iteration. The decision upon selected $ a $ is made in \textit{switching logic} sub-block, which is an essential part of the described \textit{controller}.\

	The \textit{switching logic} requires both controllers' actions and the state of the plant as the inputs. Depending on the $ s $ it decides, which one of the two control actions will proceed to the Plant. The block prioritizes $ a^{learn} $ every time, when it does not conflict with the $ a^{safe} $, because according to the strategy described in the next Section, the reward function also includes a penalty for Safety controller activation. Therefore, in short, the \textit{switching logic} allows the Learned controller for learning a policy on how to select an action, which does not lead to Safety controller activation.\
	
	Ultimately, despite its utilization during the training procedure, the \textit{switching logic} remains effective during the \textit{controller} exploitation. This is important in case when the Learned controller transmits $ a^{learn} $, which is not in compliance with required safety on current $ s $. This may happen, when, e.g., the \textit{controller} observes from the Plant a completely unfamiliar state. Thus, it does not secure safe $ a $. In this case, the \textit{switching logic} will simply select $ a^{safe} $. If the \textit{learning algorithm} remains in the loop, the Learned controller may be, therefore, adjusted accordingly to the new observed state. This, in turn, allows the Learned controller to act as a "black-box" in the system, because, when the trained policy does not meet the safety expectations, the \textit{switching logic} simply uses the Safety controller.\
	
\subsection{Learning Algorithm}
	
	The task of the \textit{learning algorithm} is to re-shape the Learned controller to make it perform in accordance to desired specifications. For example, in RL, these specifications are defined as a combination of reward and penalty functions or other similar constraints. Hence, the \textit{learning algorithm} inputs state and additional feedback from \textit{switching logic} block $ s^{*} $. The output signal is cumulative reward $ r $, which plays a major role in the Learned controller training.\
	
	The reward evaluates the state and informs the Learned controller about the results of its action. In the framework of the RL, the Learned controller tries to maximize its reward via sampling different actions. Ideally, the Reward function contains safe, comfortable, efficient, and other performance requirements concurrently.\\
		
\section{Case Study}

\begin{figure}[!t]
	\centering
	\includegraphics[width=3.4in]{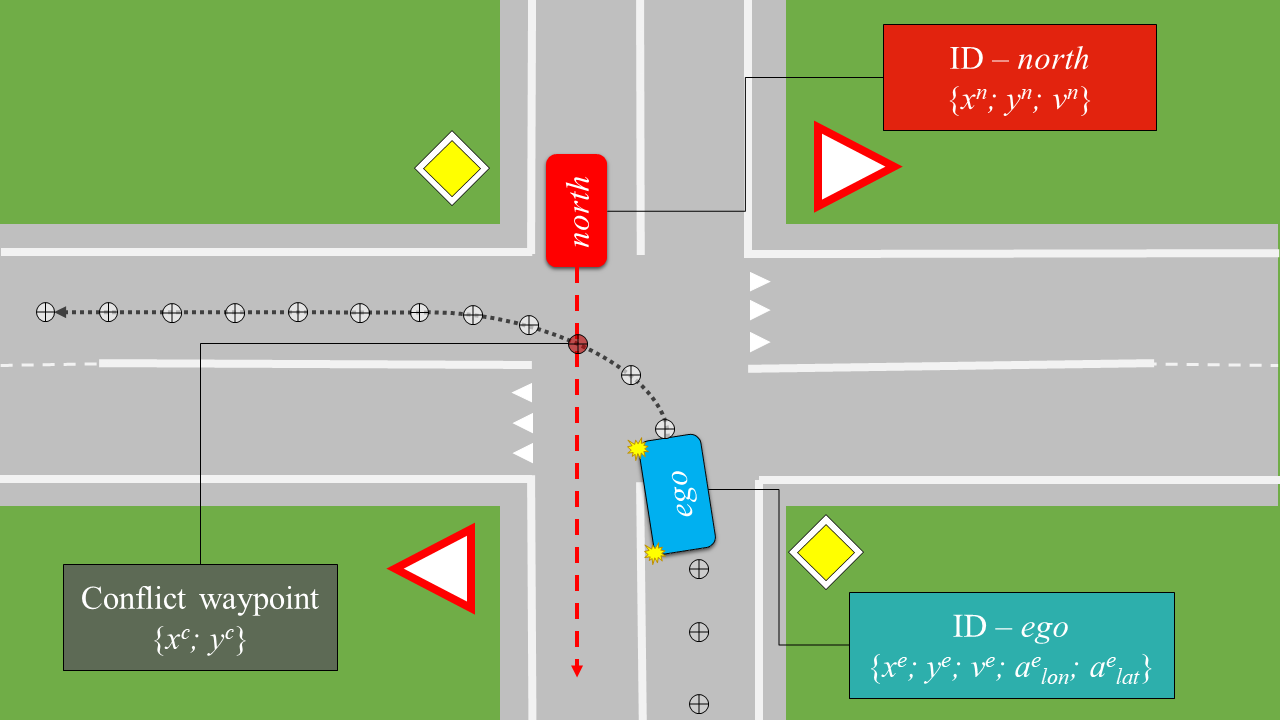}
	\caption{Case study scene: unprotected left turn.}
	\label{fig_2}
\end{figure}

	In this paper, a safety-critical maneuver of an AV is studied on proposed decision making and control system application. The case study scene is introduced in Fig.~\ref{fig_2}. The \textit{ego} (approaching from the south) is an AV. It attends to perform an unprotected left turn maneuver on cross-intersection with a mixed traffic environment, where no vehicle-to-vehicle communication is expected. It interacts with a non-AV arriving from norther leg of the intersection (the non-AV will be called \textit{north}).\	
	
	According to the road segment traffic rules, the \textit{north} has a passing priority as it drives straight. Thus, the AV must yield to the non-AV. However, it is not the best option to stop in the middle of the road waiting for the \textit{north} to clear out the path. Also, it is not the most efficient way to wait before the entrance of the intersection, and enter it only when there are no other conflicting vehicles approaching from the other legs.\
	
	Therefore, an optimal way to control the motion of the \textit{ego} AV is to apply an appropriate speed to pass the intersection not going too slow to stop nor too fast to conflict with the path of the non-AV. Furthermore, both the \textit{north} and the \textit{ego} arrive at the intersection in different moments (i.e., various $ s $). From the AV point of view, the \textit{north} may appear earlier or later: the decision making becomes multi-task. It means that on some states \textit{ego} shall yield to the \textit{north}, on other states, when there is enough time before the \textit{north} and it is safe, it can simply pass in front of the \textit{north}.\
	
	In short, a number of requirements is set to agent performance:
	
\begin{itemize}
	\item maintain safety at all costs, i.e., avoid collisions (safety);
	\item pass the intersection as fast as possible and if capable avoid stopping (efficiency);
	\item maintain desired accelerations, i.e., both lateral and longitudinal (comfort);
	\item follow a desired trajectory (safety, efficiency).
\end{itemize}

	Contemporaneously fulfilling all of the listed requirements makes the problem very complicated and challenging to solve it only with a list of well defined rules. At the same time, they can be easily expressed as separate cost or reward functions in RL. And the \textit{learning algorithm} converges the ANN in compliance with the necessary demands.\
	
\begin{figure}[!t]
	\centering
	\includegraphics[width=3.4in]{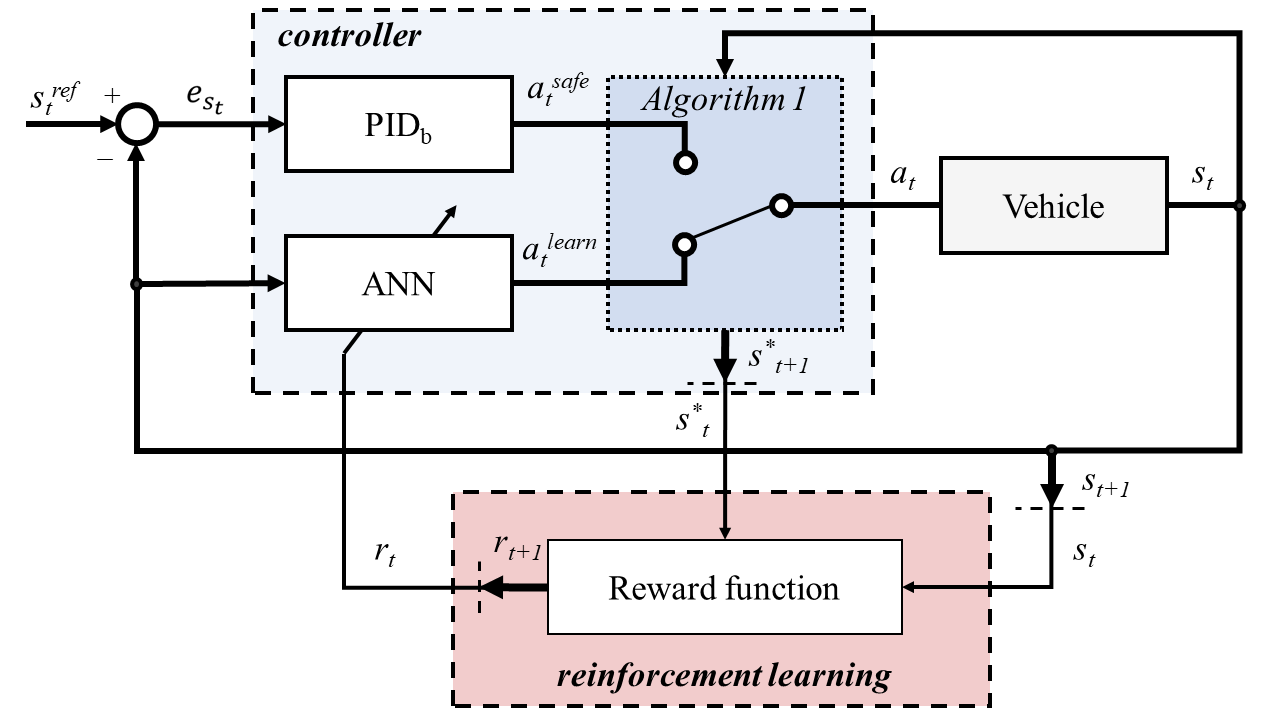}
	\caption{Decision making and control system block scheme: rule-based emergency braking ADAS and RL-based trained ANN.}
	\label{fig_3}
\end{figure}

	The case study control block scheme, which is compiled of RL as the \textit{learning algorithm} and Proportional-Integral-Derivative (PID) and ANN combination as the \textit{controller} is presented in Fig.~\ref{fig_3}. The \textit{switching logic} is an emergency braking Advanced Driver Assistance System (ADAS). It is designed with a set of rigorously defined rules.\
	
\subsection{Controller}
	
\subsubsection{Safety controller: $ PID_{b} $}
	
	For a Safety controller a traditional PID one is used. Its task is to control braking pedal displacement. The braking controller $ PID_b $ compares actual \textit{ego} AV's longitudinal deceleration $ a^{e}_{lon} $ with desired maximum deceleration rate $ a^{max}_{lon} $, selected, e.g., by a user or vehicle manufacturer. The action signal is normalized between [0 1].\
	
	The PID control rule is calculated as follows:
	
\begin{equation} \label{eq1}
	u_{t} = k_pe_{t}+k_i\int{e_{t}~dt}+k_d\frac{d}{dt}e_{t},
\end{equation}
	where $u$ is output, $e$ and $de/dt$ are input error and its time derivative, $k_p$, $k_i$, and $k_d$ - proportional, integral, and derivative coefficients, respectively. Simulation frequency $dt$ is 50 Hz.\
	
	Apart from braking controller, the \textit{ego} also accommodates throttle $ PID_{th} $ and Lane-Keeping Ability (LKA) $ PID_{lka} $ controllers. The first one tends to accelerate the AV to reference speed limit $ v^{lim} $, which is defined by the road segment. The second $ PID_{lka} $ takes care of defined trajectory following via manipulating a steering wheel. In this paper, a search algorithm $A^{*}$ is used to generate a set of waypoints the AV follows to reach its final destination. Selected PID gains, error variables, and control signal bounds are listed in Table ~\ref{tab:PIDs}.\
	
\begin{table}[!t]
	\renewcommand{\arraystretch}{1.3}
	\caption{PID controllers}
	\label{tab:PIDs}
	\centering
	\begin{tabular}{c|c|c|c|c|c}
		\hline
		\hline
		Controller		&	$e_s$						&$k_p$		&$k_i$		&$k_d$		& control signal\\
		\hline	
		$PID_{b}$		&	$ a^{max}_{lon} - a^{e}_{lon}$		&0.3			&0.40		&0.100		&[0.0 1.0] \\
		$PID_{th}$		&	$ v^{lim} - v^{e}$				&1.5			&0.05		&0.002 		&[0.0 0.7] \\
		$PID_{lka}$	&	$ lka $						&0.2			&0.01		&0.020		& [-1.0 1.0]\\
		\hline
		\hline
	\end{tabular}
\end{table}

\subsubsection{Learned controller: ANN}
	
	The Learned controller is an ANN, which is shaped during the training phase by the \textit{learning algorithm}. Ideally the AI-based model learns multi-task action $ a^{learn} $, yield or cut, depending on the time step the \textit{north} approaches the intersection. Hence, the ANN must be large enough to guarantee its multi-task response.\
	
	After several experiments, the size of the feedforward ANN was established with three hidden layers of 128 neurons in each one. The activation function of the input is rectified linear unit. The ANN input layer receives the state $ s $ directly from the Plant (i.e., \textit{ego} AV), and its continuous output layer's control signal is a throttle pedal displacement normalized inside [0 1]. The activation function of the output layer is $ tanh $.\
		
\subsubsection{Switching logic: emergency braking ADAS}
	
\begin{algorithm}[!t]
\KwData{ego AV's velocity $v^e$ and location ($x^e$, $y^e$); \textit{north} vehicle's velocity $v^n$ and location ($x^n$, $y^n$); conflict waypoint coordinates ($ x^{c} $, $ y^{c} $); ego AV's route waypoints; intersection geometry and traffic rules including speed limt $ v^{lim} $; AV's final velocity $ v_f $ (0 at full stop); maximum longitudinal acceleration $ a_{lon}^{max} $, AV's lane-keeping ability $ lka $; control signals from Safe ($ PID_b $) and Learned (ANN) controllers $a^{safe}$ and $a^{learn}$, correspondingly.\

	\textit{Note}: superscripts $e$ and $n$ refer to \textit{ego} and \textit{north} vehicles, accordingly.}
\KwResult{Longitudinal behaviour (braking or accelerating) for AV}
$steer^{e}\gets PID_{lka}^{e}(lka)$\;
$throttle^{e}\gets PID_{th}^{e}( v^{lim} - v^{e})$\;
$brake^{e}\gets 0$\;
\While{$ ego $ and $ north $ are both at intersection}{
	\For{each time step $t$}{
		$TTR^{n}_{t}\gets Eq.~\ref{eq2}$\;
		$TTR^{e}_{t}\gets Eq.~\ref{eq2}$\;
		\eIf{$ TTR^{n}_{t}-TTR^{e}_{t}>threshold $}{
			$v_{i_{t}}\gets v_{t}^{e}$\;
			$d_{safe_{t}}^{e}\gets Eq.~\ref{eq3}$\;
			\eIf{$ d_{safe_{t}}< d_{c_{t}}^{e} $ }{
				$a_{t}\gets a_{t}^{safe}$\;}
			{$a_{t}\gets a_{t}^{learn}$\;}
		}
		{$a_{t}\gets a_{t}^{learn}$\;}
	}
}
\caption{Rule-based emergency braking ADAS}
\label{alg:alg_1}
\end{algorithm}
	
	The core element of the \textit{controller}  is the \textit{switching logic} (Fig.~\ref{fig_1}). For the case study, the goal of the \textit{switching logic} is to select a safe action $ a^{safe} $, which ensures collision avoidance with the \textit{north} non-AV. The safe action must be only applied, when the algorithm identifies that under current state the collision with other non-AV is inevitable.\
	
	During unprotected left turn exists a potential conflict point, where collision may happen between the \textit{ego} and the \textit{north}. This waypoint ($ x^{c} $, $ y^{c} $) is highlighted with red color in Fig.~\ref{fig_2}. This conflict location is extracted from high definition maps whenever two vehicles are meeting at an intersection. When two car cross an intersection they always have at least one location, where they share the same waypoint from their globally planned trajectories.\
	
	As the routes of the vehicles inter-cross at the intersection in a given case study, the conflict waypoint belongs to the list of waypoints the AV and the \textit{north} follow. It is an approximate location, where the \textit{ego's} path conflicts with the \textit{north's} one, if both vehicles follow the traffic rules and remain in their driving lanes. The conflict waypoint is extracted from a static information about the intersection (i.e., shape, type, rules, etc.). The vehicles must not meet at this location, thus, if necessary, the \textit{ego} must stop before the conflict waypoint to yield to the \textit{north}, or pass before the \textit{north} reaches it. In this case study, the \textit{switching logic} takes form of an emergency braking ADAS presented in Algorithm~\ref{alg:alg_1}.\ 
	
	When the AV enters an intersection, the time-to-reach (TTR) variable is calculated for both \textit{ego} and \textit{north}, which depends on their location and speed they are traveling with. The TTR is widely used for maneuver safety assessment~\cite{Nayak17}, in vehicle safety systems like adaptive cruise control and collision avoidance. This index allows for estimating time when two moving objects will collide given their routs intersect. The simplified way to calculate TTR is as follows:

\begin{equation} \label{eq2}
	TTR = \frac{d_c}{v},
\end{equation}
where $ d_c $ is a distance to conflict waypoint (red point in Fig.~\ref{fig_2}) from vehicle's current location, $ v $ is car's velocity.\
	
	The distance $ d_c $ is calculated as a summations of the Euclidean distance formulations between proceeding waypoints. Therefore, it is also applicable for the \textit{ego} that travels curved trajectory during the left turn.\
	
	In general, the TTR index points at time period a vehicle will reach a specific location at its route with a given speed. When two vehicles has a potential to collide due to the inter-cross of the trajectories, the TTR can be used to asses, how much time does each car need to reach the conflict location? Hence, the difference between two values (i.e., difference between the TTR's of the \textit{ego} and the \textit{north}) can be used to evaluate probable collision at given speeds. A \textit{threshold} is introduced to evaluate, if the AV has enough time to cut in front of the non-AV.\
	
	The emergency braking ADAS ensures that the AV will stop before the waypoint prior the conflict one by applying safe action, when cutting is not an option. For this, it calculates the minimum safe distance $ d_{safe} $, from where the AV must always switch to the Safe controller and apply the $ a^{safe} $ to ensure braking at maximum desired rate and not further away from the waypoint located right before the conflict one. The following formula is used: 
	
\begin{equation} \label{eq3}
	d_{safe} = \frac{v^{2}_{f} - v^{2}_{i}}{2a_{lon}^{max}},
\end{equation}
where $ a_{lon}^{max} $ is a maximum longitudinal acceleration rate, $ v_i $ and $ v_f $ are AV's initial and final velocities, respectfully. Initial velocity is \textit{ego's} speed $ v^e $ in Algorithm~\ref{alg:alg_1}.\	
			
	The minimum safe distance is then extracted to the sum of the Euclidean distances between proceeding waypoints of the \textit{ego's} route, what enables to account for curved trajectory of the AV. As a result, the \textit{ego} knows exact location on its path from where to apply brakes for stopping in front of the conflict waypoint at the intersection. For instance, assume that $ s^{safe} $ is 10 m, and the distance between each AV's trajectory waypoints is 2 m. Hence, neglecting lateral dynamics of the vehicle, the \textit{ego} will apply emergency braking 5 waypoints away from the desired stopping location (e.g., 1 waypoint before the conflicting one).\
	
	Next, the distance between current location of the \textit{ego} and the location of the conflict waypoint at the intersection $ d_c^{e} $ as a summation of Euclidean distances between proceeding waypoints in front of the AV is compared to the calculated safe distance $ d_{safe} $. Finally, when Algorithm~\ref{alg:alg_1} understands that $ d_{safe} $ is smaller than $ d_c $ (collision is unavoidable without braking while conflicting vehicles are both at the intersection), it applies emergency braking action $ a^{safe} $ to the AV, otherwise - the Learned controller action $ a^{learn} $. Algorithm~\ref{alg:alg_1} runs at each time step $ t $ while both vehicles are at the intersection. Final control action $ a $ is sent to the Plant.\

\subsection{Learning Algorithm}
	
\begin{figure}[!t]
	\centering
	\subfloat[]{\includegraphics[width=3.8in]{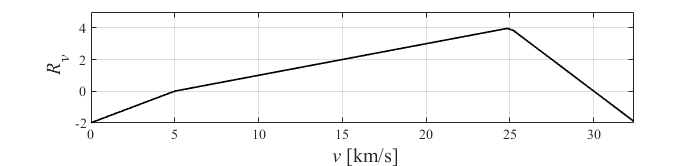}%
		\label{fig_4_a}}
	\label{fig_4a}
	\subfloat[]{\includegraphics[width=3.8in]{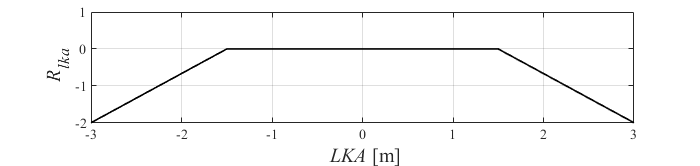}%
		\label{fig_4_b}}
	\label{fig_4b}
	\subfloat[]{\includegraphics[width=3.8in]{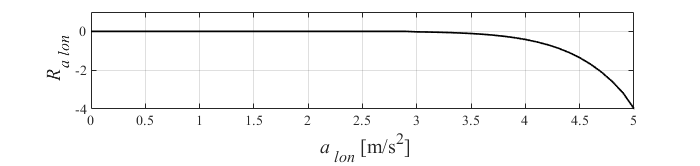}%
		\label{fig_4_c}}
	\label{fig_4c}
	\subfloat[]{\includegraphics[width=3.8in]{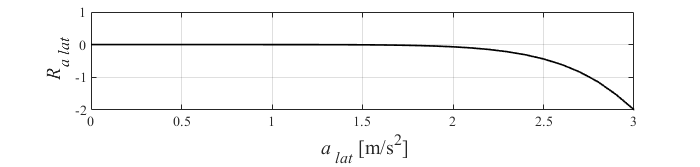}%
		\label{fig_4_d}}
	\label{fig_4d}
	\caption{Reward functions: (a) velocity (efficiency); (b) lane-keeping ability (safety); (c) longitudinal acceleration (comfort); (d) lateral acceleration (comfort).}
	\label{fig_4}
\end{figure}
		
	The \textit{learning algorithm} is based on RL and utilizes the Proximal Policy Optimization (PPO) algorithm. The PPO is one of the most popular RL algorithms, because it proved to be robust to hyperparameters' initialization. It is a policy gradient method that updates policy using a surrogate loss function to avoid significant performance decrease~\cite{Schulman17}.\
	
	One of the main steps in every RL algorithm is a design of reward functions. While, for some problems the default hyperparameters may be applied, the user still must adjust or design from scratch the reward functions. Moreover, if the problem to be solved is noticeably complex, meaning that multiple requirements must be taken into consideration, the separate reward functions play major roles in successful convergence of the policy.\
	
	As it is stated earlier in this section, intersection crossing is a very complex problem, because it has a number of requirements to be accomplished. Namely, maximizing AV's speed points at efficiency, because decreasing speed may be considered as energy loss due to friction forces between tires and road surface. Not activating the Safety controller and maintaining the trajectory line (precisely following the waypoints) jointly indicate safety. Lastly, keeping the lateral and longitudinal accelerations below desired limits is responsible for vehicle occupants' comfort and motion sickness.\
	
\subsubsection{State}
	
	The state $ s $ consists of a number of variables measured by the available on-board sensor set. It includes velocities $ v $ and $ x $, $ y $ positions of both the \textit{ego} and the \textit{north} vehicles: $ s = (x^e, y^e, v^e, x^n, y^n, v^n) $. The state is also an input layer to the ANN.\
	
	\textit{Remark 2:} Perception and localization of the \textit{ego} AV are not part of this work nor part of the state. Nevertheless, the AV must first recognize other dynamic traffic participants before estimating their relative locations and speeds.\
	
\subsubsection{Reward}
	
	The emergency braking ADAS activation plays a principal role in reward function. Hence, the \textit{switching logic} block outputs additional state $ s^* $, which is not a part of the main state $ s $ transmitted from the Plant. The ADAS reward $ R_{ADAS} $ gives a penalty for activation of the Safety controller:
	
\begin{equation} \label{eq4}
	R_{ADAS_{t}} = 
	\begin{cases} 
	-25 & \text{if } a_{t} = a_{t}^{safe}\\
	0    & \text{if } a_{t} \neq a_{t}^{safe}
	\end{cases}
\end{equation}

	All the rest of the rewards related to velocity $ R_{v} $, LKA $ R_{lka} $, longitudinal $ R_{a_{lon}} $ and lateral $ R_{a_{lat}} $ accelerations are drawn in Fig.~\ref{fig_4}. The final cumulative reward $ r $ is expressed as a summation of all five rewards:
	
\begin{equation} \label{eq5}
	r_{t} = R_{ADAS_{t}}+R_{lka_{t}}+R_{v_{t}}+R_{a_{lon_{t}}}+R_{a_{lat_{t}}}
\end{equation}
	
	The AV velocity reward (Fig.~\ref{fig_4_a}) requires a speed limit of the road section. In this case study, the value is limited to 25 km/h. The maximum reward is given, when the AV reaches it. In addition, to motivate the AV to move as fast as feasible, yet not exceeding the speed limit, the reward decreases with the velocity descent. Furthermore, the reward also outputs penalty for very low speeds, below 5 km/h in this design. Finally, the reward allows speeding, but only up to 5 km/h higher than road speed limit. The AV's speed over 35 km/h (speed limit plus allowed speeding) gives a penalty.\
	
	Another reward that is responsible for safety is LKA (Fig.~\ref{fig_4_b}). It allows for 1.5 meters of AV's center deviation from the middle line of its trajectory to the left or to the right sides. While, $ PID_{lka} $, which is not a part of the Learned controller, is responsible for LKA maneuver, the action of the controller $ a^{learn} $ directly affects the stability of lateral control. Hence, the LKA reward sightly contributes to ensuring that the ANN will find an action, which will not negatively impact the lateral control of the car (i.e., the stability will be preserved).\
	
	At last, the rewards responsible for the comfort of the maneuver, longitudinal (Fig.~\ref{fig_4_c}) and lateral (Fig.~\ref{fig_4_d}) accelerations are proposed. They are limited to the general guidance of passenger vehicle comfort~\cite{ECE20}, namely 5 m/s$^{2}$ and 3 m/s$^{2}$ for longitudinal and later accelerations, respectively.\\
	
\section{Training Results}
	
\subsection{Simulation Environment}
		
\begin{table}
	\renewcommand{\arraystretch}{1.3}
	\caption{PPO Hyperparameters}
	\label{tab:hyperparameters}
	\centering
	\begin{tabular}{c|c}
		\hline
		\hline
		Parameter							&  Value\\
		\hline	
		Discount factor						&   0.99\\			
		Horizon							&   128\\
		Entropy coefficient					&   0.01\\			
		Learning rate    						&   0.00025\\
		Value function coefficient   				&   0.5\\			
		Maximum value for the gradient clipping    	&   0.5\\
		Clip range   						&   0.2\\
		Lambda							&	0.9\\
		Number of training minibatches			&	4\\
		Number of epoch					&	4\\
		Time steps							&   1 200 000\\
		Number of hidden units per layer			&	[128, 128, 128]\\
		Optimizer							&	Adam\\
		Normalization						&	Observation and reward\\
		Non-linearity						&	$ Tanh $\\
		\hline
		\hline
	\end{tabular}
\end{table}
		
\begin{figure}[!t]
	\centering
	\subfloat[]{\includegraphics[width=3.8in]{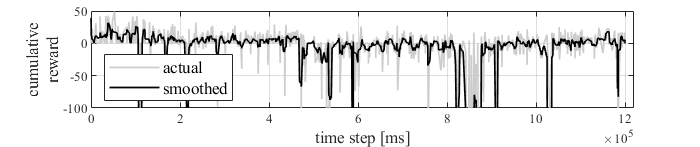}%
		\label{fig_5_a}}
	\label{fig_5a}
	\subfloat[]{\includegraphics[width=3.8in]{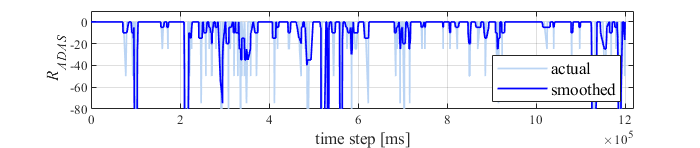}%
		\label{fig_5_b}}
	\label{fig_5b}
	\subfloat[]{\includegraphics[width=3.8in]{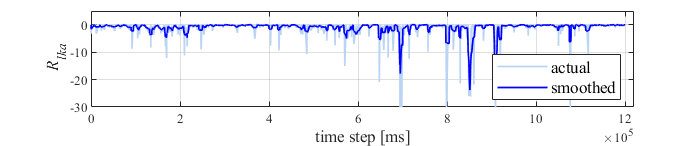}%
		\label{fig_5_c}}
	\label{fig_5c}
	\subfloat[]{\includegraphics[width=3.8in]{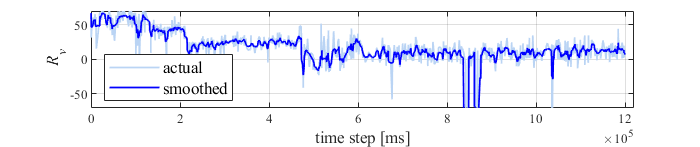}%
		\label{fig_5_d}}
	\label{fig_5d}
	\subfloat[]{\includegraphics[width=3.8in]{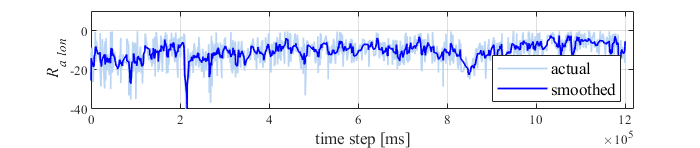}%
		\label{fig_5_e}}
	\label{fig_5e}
	\subfloat[]{\includegraphics[width=3.8in]{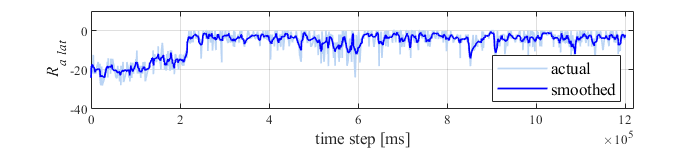}%
		\label{fig_5_f}}
	\label{fig_5f}
	\caption{Rewards: (a) cumulative (non-normalized); (b) ADAS; (c) lane-keeping ability; (d) velocity; (e) longitudinal acceleration; (f) lateral acceleration.}
	\label{fig_5}
\end{figure}
	
	For the case study of the decision making and control system an open source CARLA simulator (0.9.5)~\cite{Dosovitskiy17} running on Linux operating system was used. The simulator allows for testing various complex scenes under different weather and road conditions. The simulator appeared to be applicable for AV algorithms numerical validation. It has a client-server architecture, whereas the client is implemented in Python programming language. The motion of the vehicle is regulated applying steering, acceleration, and braking commands.\
	
	The RL-based ANN is trained with Stable Baselines open source library~\cite{Hill18} created in the robotics lab U2IS (INRIA Flowers team) at ENSTA-ParisTech (Paris, France). The library allows for utilizing state-of-the-art RL algorithms directly in custom environment with custom or default policies. In this case study, the RL algorithm PPO was applied due to its simplicity and effectiveness in solving various complex control and decision making problems~\cite{Schulman17}.\
	
	The hyperparameters selection is very important for training efficacy for every RL algorithm. For this, the recommendations from~\cite{Andrychowicz20} were carefully studied. Indicatively, the chosen values for PPO algorithm along with the ANN hyperparameters are listed in Table~\ref{tab:hyperparameters}.\
	
\subsection{Training Results}
	
	The training results of the cumulative reward and individual rewards are demonstrated in Fig.~\ref{fig_5}. The non-normalized cumulative reward (Fig.~\ref{fig_5_a}) is the main criteria for learned model validation. Ideally, during successful learning it remains as large as possible. However, sometimes the ANN explores new actions that lead to Safety controller activation, and the cumulative value drops down noticeably.\
	
	As the ADAS reward (Fig.~\ref{fig_5_b}) is a penalty for activating the emergency braking system, the algorithm is motivated not to involve it. During the first 220 000 ms the RL explores on AV's speed maximization (Fig.~\ref{fig_5_d}). However, lateral acceleration reward dramatically decreases in these cases (Fig.~\ref{fig_5_f}). Hence, the algorithm finds optimal solution to satisfy both requirements. Finally, by the end of the training, the learned behavior minimizes the mistakes made by LKA (Fig.~\ref{fig_5_c}), reduces the influence of lateral (Fig.~\ref{fig_5_f}) and longitudinal (Fig.~\ref{fig_5_e}) accelerations, and identifies the maximum possible velocity (Fig.~\ref{fig_5_d}) of the AV.\\

\section{Evaluation Results}

	In the case study, the AV learns multi-task policy. Conditioned by the input state, it selects a policy action either to cut in front of the \textit{north} or yield to it. The \textit{north} drives in the \textit{autopilot} mode embedded in the simulator. In this mode, however, the function of yielding to other vehicles is turned off. To demonstrate robustness of the multi-task ANN, the results for different states are presented next.\
	
	\textit{Remark 3:} The lateral control (i.e., LKA and steering wheel control signal from $PID_{lka}^{e} $) is not submitted in this paper, because it is not a part of the examined decision making and control system.\

\subsection{Policy 1 - Cutting}
		
\begin{figure}[!t]
	\centering
	\subfloat[]{\includegraphics[width=3.8in]{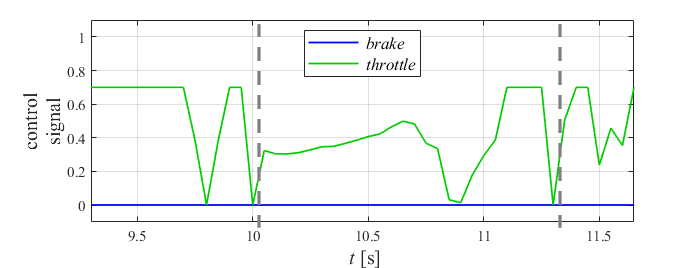}%
		\label{fig_8_a}}
	\label{fig_8a}
	\subfloat[]{\includegraphics[width=3.8in]{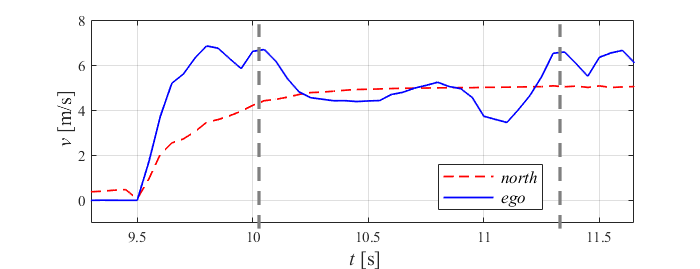}%
		\label{fig_8_c}}
	\label{fig_8c}
	\subfloat[]{\includegraphics[width=3.8in]{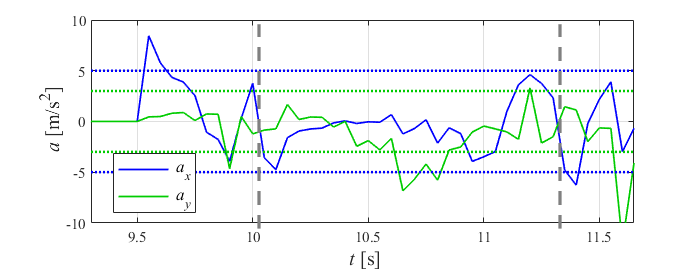}%
		\label{fig_8_d}}
	\label{fig_8d}
	\caption{Experimental results on cutting in front of the \textit{north} car: (a) control signals; (b) speeds; (c) accelerations (the horizontal dashed lines emphasize the training constraints for longitudinal and lateral accelerations). The vertical dashed lines highlight the period, when the AV is at the intersection.}
	\label{fig_8}
\end{figure}

	The experimental results of the \textit{ego} AV cutting in front of the \textit{north} non-AV are shown in Fig.~\ref{fig_8}. The \textit{north} appears late at the intersection, what gives the AV an opportunity not to wait, and safely cross the intersection in front of the non-AV.\
	
	In Fig.~\ref{fig_8_a}, the control signals (i.e., actions $ a^{learn} $ - throttle and $ a^{safe} $ - brake) are drawn. It is clearly seen that the brake signal, the $ a^{safe} $ delivered from $ PID_{b} $, is not activated during the whole maneuver, symbolizing that the ANN learned the policy action, which avoids the situations, where the emergency braking ADAS is switched on. Furthermore, the output of the Learned controller is a smooth continuous action controlling the throttle pedal.\
	
	In Fig.~\ref{fig_8_c}, the vehicles' velocities are presented. The \textit{north} vehicle travels without significant speed change. It symbolizes that it did not need to stop in front of the \textit{ego} to let it drive ahead. Despite it was demanded by the reward function (Fig.~\ref{fig_4_a}), the velocity of the \textit{ego} AV is much below the speed limit of the road segment (i.e., 6.9 m/s or 25 km/h). This is because during the optimization the AV looks for maximization of a cumulative reward. Because it is not always feasible to reach the maximum possible cumulative reward, the \textit{ego} sacrifices velocity in order to fulfill the requirements set by other training constraints, namely activation of the Safety controller or lateral and longitudinal accelerations.\
		
	In addition, although the throttle pedal is positive and the braking one - zero (Fig.~\ref{fig_8_a}), the velocity of the AV slightly decreases during a short period at the intersection. This is due to vehicle dynamics model embedded in the simulator, namely the requested driving torque (i.e., throttle pedal) is not enough to overcome the friction forces, and, thus, the velocity of the AV decreases. Though, the trained Agent accounts for the vehicle dynamics without knowing precise plant model.\
	
	The longitudinal and lateral accelerations of the \textit{ego} vehicle are plotted in Fig.~\ref{fig_8_d}. In accordance to the set requirements (i.e., 5 m/s$^2$ for longitudinal and 3 m/s$^2$ for lateral) the ANN tries to keep the AV inside of them. While the performance of the AV for the longitudinal acceleration is within the acceptable range, the lateral counterpart still scarcely valuates its requirements boundaries.\

\subsection{Policy 2 - Yielding}
			
\begin{figure}[!t]
	\centering
	\subfloat[]{\includegraphics[width=3.8in]{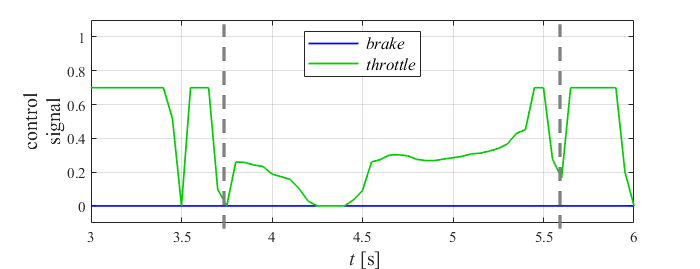}%
		\label{fig_9_a}}
	\label{fig_9a}
	\subfloat[]{\includegraphics[width=3.8in]{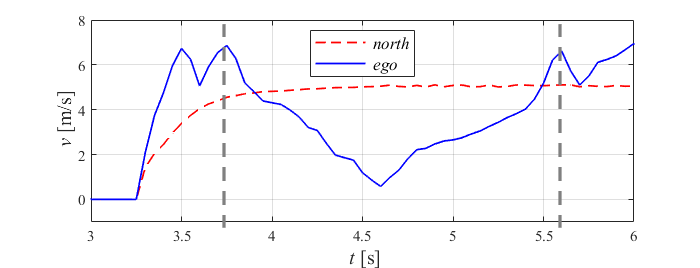}%
		\label{fig_9_c}}
	\label{fig_9c}
	\subfloat[]{\includegraphics[width=3.8in]{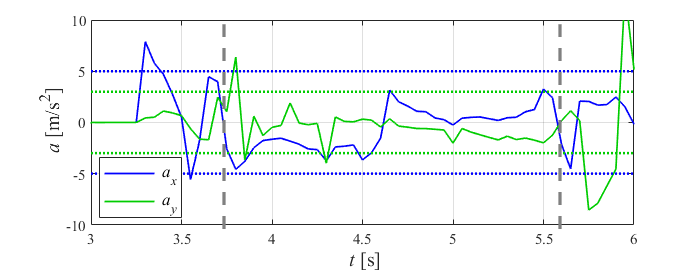}%
		\label{fig_9_d}}
	\label{fig_9d}
	\caption{Experimental results on yielding to the \textit{north} car state 1: (a) control signals; (b) speeds; (c) accelerations (the horizontal dashed lines emphasize the training constraints for longitudinal and lateral accelerations). The vertical dashed lines highlight the period, when the AV is at the intersection.}
	\label{fig_9}
\end{figure}
	
\begin{figure}[!t]
	\centering
	\subfloat[]{\includegraphics[width=3.8in]{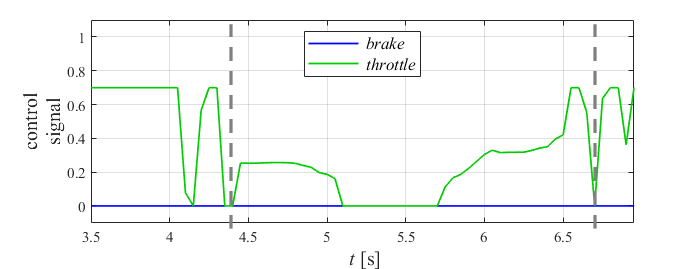}%
		\label{fig_10_a}}
	\label{fig_10a}
	\subfloat[]{\includegraphics[width=3.8in]{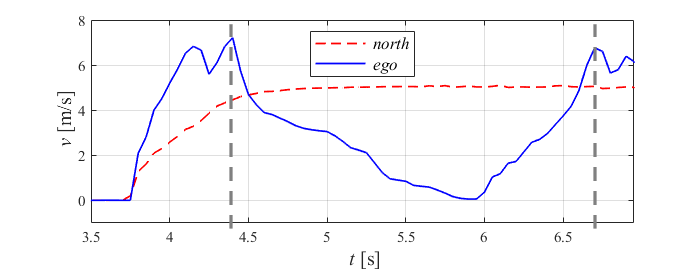}%
		\label{fig_10_c}}
	\label{fig_10c}
	\subfloat[]{\includegraphics[width=3.8in]{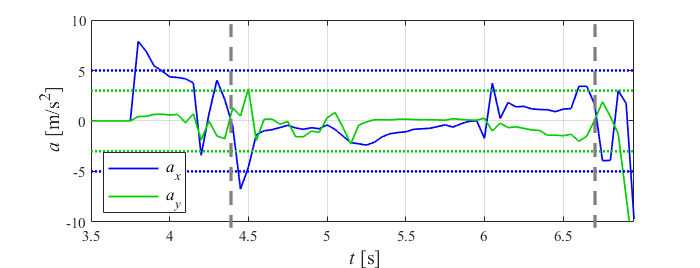}%
		\label{fig_10_d}}
	\label{fig_10d}
	\caption{Experimental results on yielding to the \textit{north} car state 2: (a) control signals; (b) speeds; (c) accelerations (the horizontal dashed lines emphasize the training constraints for longitudinal and lateral accelerations). The vertical dashed lines highlight the period, when the AV is at the intersection.}
	\label{fig_10}
\end{figure}
	
\begin{figure}[!t]
	\centering
	\subfloat[]{\includegraphics[width=3.8in]{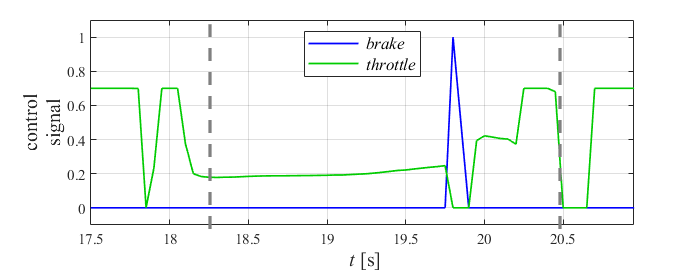}%
		\label{fig_11_a}}
	\label{fig_11a}
	\subfloat[]{\includegraphics[width=3.8in]{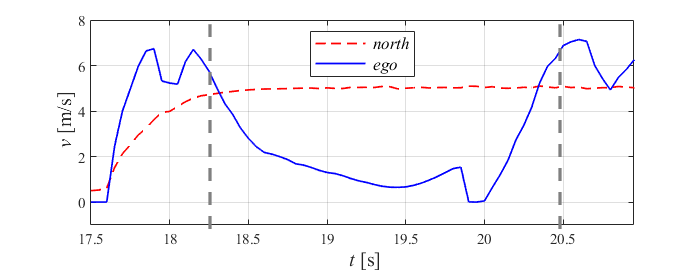}%
		\label{fig_11_c}}
	\label{fig_11c}
	\subfloat[]{\includegraphics[width=3.8in]{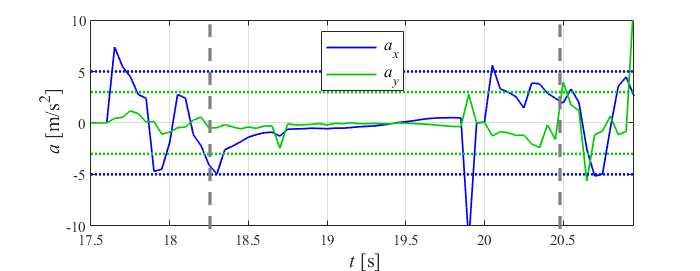}%
		\label{fig_11_d}}
	\label{fig_11d}
	\caption{Experimental results on trained ANN with removed $ R_{ADAS} $: (a) control signals; (b) speeds; (c) accelerations (the horizontal dashed lines emphasize the training constraints for longitudinal and lateral accelerations). The vertical dashed lines highlight the period, when the AV is at the intersection.}
	\label{fig_11}
\end{figure}

	For additional method's robustness demonstration two cases on yielding to the \textit{north} are reported. It is essential for revealing the influence of state differences on ANN's response. Namely, the non-AV arrives at the intersection at different times, however, in both cases the AV does not have enough time to safely cut in front of the \textit{north}. Therefore, it must adjust the speed and preferably not to fully stop in the middle of the intersection as it is demanded by the reward function during the training stage.\
	
	For robustness two cases are presented in Fig.~\ref{fig_9} and Fig.~\ref{fig_10}. Again, it can be noticed that in both runs, the braking signal coming from the Safety controller was not requested (Fig.~\ref{fig_9_a} and Fig.~\ref{fig_10_a}). Moreover, the throttle control signal from the learned ANN is smooth and continuous.\
	
	The velocity plots are displayed in Fig.~\ref{fig_9_c} and Fig.~\ref{fig_10_c}. The differences between two cases are recognized here. While the \textit{north} maintains its speed in both cases, the \textit{ego} is able to cross the intersection without a full stop in the first case study (Fig.~\ref{fig_9_c}). However, in the second demonstration (Fig.~\ref{fig_10_c}), the AV drops the speed to a full stop for a few milliseconds allowing the non-AV to pass and then immediately accelerates to the desired speed. This shows that the \textit{north} arrives at the intersection later than in the first case: it demands from the AV to drive at the intersection slower, up until a full stop for a negligibly small period of time.\
	
	Lastly, the longitudinal and lateral acceleration rates prove the smooth and comfortable maneuver for the passengers of the AV in studied examples (Fig.~\ref{fig_9_d} and Fig.~\ref{fig_10_d}). Both variables are inside the required limits leading to protection against the aggressive performance, and lane following stability maintenance.\
	
	In short, both cutting and yielding operations demonstrate ANN's ability to learn multi-task policy. Furthermore, the simulation results illustrate the proposed system's ability to satisfy multiple requirements set by the user, e.g., safety, efficiency, and comfort. Hence, the provided method is especially useful in solving complex multi-task decision making and control challenges.\
	
	\textit{Remark 4:} As it is suggested in~\cite{Hill18}, a longer training may lead to more accurate policy that even better satisfies multiple requirements, e.g., lateral acceleration in cutting scenario. The purpose of this paper, however, is only a proof of concept of the proposed decision making and control framework, where detailed study on hyperparameters adjustment in the \textit{learning algorithm} is out of scope in current study.\
	
\subsection{\textit{Switching Logic} Impact Demonstration}
	
	To demonstrate the positive impact of the \textit{switching logic} participation in the Learned controller training, one additional ANN model was trained. The emergency braking ADAS reward $ R_{ADAS} $ transmitted from the \textit{switching logic} was removed from Eq.~\ref{eq5}. For comparative analysis purpose the results are delivered in Fig.~\ref{fig_11}. For proper analysis, the yielding case is addressed, because in the cutting in front of the \textit{north} scene the activation of the ADAS is not guaranteed.\
	
	The Learned controller indeed finds a continuous policy for the Plant control. It means that the converged model satisfied all requirements except the main safety one. However, it is still guilty for braking pedal activation while yielding to the non-AV (Fig.~\ref{fig_11_a}). This cases study simulates the failure of the Learned controller when the RB Safety controller steps in to avoid traffic accident.Braking pedal actuation request shows that \textit{switching logic} passed the signal transmitted from Safety controller, and during this period the Learned counterpart was deactivated. Hence, thanks to the robust RB function, safety is presumed. Though, the vehicle comes to a full stop (Fig.~\ref{fig_11_c}), but negatively impacts motion efficiency losses, not mentioning the activation of the braking system that increases brake particle emission. Furthermore, the comfort of the AV is significantly reduced due to the activation of the brakes (Fig.~\ref{fig_11_d}).\
	
	Nevertheless, in comparison to the yielding results of the first model (Fig.~\ref{fig_10}), where the \textit{ego} comes to a full stop, too, the activation of the brake pedal still affects the overall performance in a negative way. It does not ensure the set requirements satisfaction every time. Moreover, it always impacts on efficacy of the trip, because during braking, the vast amount of energy is wasted on heat.\
	
\subsection{Limitations}
	
	Few vital limitations of the developed decision making and control system must be mentioned. Firstly, the case study selected for method experimental validation is relatively simple, however, it can be re-designed for more complicated trial, e.g., for a number of non-AVs arriving from each separate leg of the intersection. Secondly, for successful implementation of the proposed solution, the developer depends on accurate Safety controller and \textit{switching logic} design. Ultimately, although the Learned controller is secured by the Safety one, it still serves as a "black-box", what makes it exposed to failing to meet some functional safety requirements in practical application.\\
		
\section{Conclusion}

	In this paper, a decision making and control framework is proposed. It combines a ML algorithm with a RB control method: the Learned and Safety controllers. The Safety controller is responsible for ensuring safety at any cost, thus, sacrificing comfort and efficiency of the Plant performance. The Learned controller aims at fulfilling multiple requirements simultaneously: comfort, efficiency, and safety. It is based on ML, because for modern control systems it is a burden to design RB controllers, which achieves scalability.\
	
	The proposed method includes a \textit{switching logic}. Referring to the state, its role is to identify safety-threatening situations and switch between Safety and Learned controllers. The Safety controller has a priority, whenever safe operation of the latter one is not certain. During the training phase the Learned controller searches for the action output, which allows for not activating the Safety controller, yet, at once satisfying other relevant constraints (e.g., comfort and efficiency). The simulation results demonstrate that the proposed method prosperously meets the established demands for AV behavior: it simultaneously maximizes efficiency, comfort, and safety. Finally, the method allows for meeting safety constraints (i.e., avoid collisions) not only during the exploitation process, but also during the ANN training.\\


\begin{thebibliography}{}

\bibitem{Michalski83}
R.S. Michalski, J.G. Carbonell, and T.M. Mitchell (Eds.), \emph{Machine Learning: An Artificial Intelligence Approach}, Berlin, Germany: Springer-Verlag Berlin Heidelberg, 1983, pp. 3-23. DOI: 10.1007/978-3-662-12405-5

\bibitem{Yoon94}
Y. Yoon, T. Guimaraes, and G. Swales, "Integrating artificial neural networks with rule-based expert systems," \emph{Decision Support Sys.}, vol. 11, no. 5, pp. 497-507, June 1994. DOI: 10.1016/0167-9236(94)90021-3

\bibitem{Likmeta20}
A. Likmeta, A.M. Metell, A. Tirinzoni, R. Giol, M. Restelli, and D. Romano, "Combining reinforcement learning with rule-based controllers for transparent and general decision-making in autonomous driving," \emph{Rob. and Auton. Sys.}, vol. 131, pp. 1-17, 2020. DOI: 10.1016/J.ROBOT.2020.103568

\bibitem{Polanyi}
M. Polanyi, \emph{Personal Knowledge: Towards a Post-Critical Philosophy}, Chicago, USA: University of Chicago Press, 1958, pp. 1-25.

\bibitem{Hewing20}
L. Hewing, K.P. Wabersich, M. Menner, and M.N. Zeilinger, "Learning-based model predictive control: Towards safe learning in control," \emph{Annu. Rev. Control Robot. Auton. Syst.}, vol. 3, pp. 269-296, May 2020. DOI: 10.1146/ANNUREV-CONTROL-090419-075625
	
\bibitem{Maravall09}
D. Maravall, J. de Lope, and J.A. Martin H., "Hybridizing evolutionary computation and reinforcement learning for the design of almost universal controllers for autonomous robots," \emph{Neurocomputing}, vol. 72, pp. 887–894, Nov. 2008. DOI: 10.1016/J.NEUCOM.2008.04.058

\bibitem{Franca20}
M.A. Franco, N. Krasnogor, and J. Bacardit, "Automatic tuning of rule-based evolutionary machine learning via problem structure identification," \emph{IEEE Comput. Intel. Magaz.}, vol. 15, no. 3, pp. 28-46, July 2020. DOI: 10.1109/MCI.2020.2998232

\bibitem{He20}
G. He, X. Xin, R. Peng, M. Han, J. Wang, and X. Wu, "Online rule-based classifier learning on dynamic unlabeled multivariate time series data," \emph{IEEE Trans. on Sys., Man, and Cyber.: Sys.}, vol. 52, no. 2, pp. 1121-1134, Aug. 2020. DOI: 10.1109/TSMC.2020.3012677

\bibitem{Wang20}
H. Wang, Y. Huang, A. Khajepour, D. Cao, and C, Lv, "Ethical decision-making platform in autonomous vehicles with Lexicographic Optimization based model predictive controller," \emph{IEEE Trans. on Veh. Tech.}, vol. 69, no. 8, pp. 8164-8175, May 2020. DOI: 10.1109/TVT.2020.2996954

\bibitem{Bhattacharyya15}
S. Bhattacharyya, D. Basu, A. Konar, and D.N. Tibarewala, "Interval type-2 fuzzy logic based multiclass ANFIS algorithm for real-time EEG based movement control of a robot arm," \emph{Robot. and Auton. Sys.}, vol. 68, pp. 104-115, June 2015. DOI: 10.1016/J.ROBOT.2015.01.007

\bibitem{Yeh11}
C.-Y. Yeh, W.-H. R. Jeng, and S.-J. Lee, "Data-based system modeling using a type-2 fuzzy neural network with a hybrid learning algorithm," \emph{IEEE Tran. on Neural Net.}, vol. 22, no. 12, pp. 2296-2309, Dec. 2011. DOI: 10.1109/TNN.2011.2170095

\bibitem{Khanesar12}
M.A. Khanesar, E. Kayacan, M. Teshnehlab, and O. Kaynak, "Extended Kalman filter based learning algorithm for type-2 fuzzy logic systems and its experimental evaluation," \emph{IEEE Tran. on Indust. Elec.}, vol. 59, no. 11, pp. 4443-4455, Nov. 2012. DOI: 10.1109/TIE.2011.2151822

\bibitem{Ibrahim20}
H.A. Ibrahim, A.T. Azar, Z.F. Ibrahim, and H.H. Ammar, "A hybrid deep learning based autonomous vehicle navigation and obstacles avoidance," in \emph{Proc. of the 2020 21\textsuperscript{st} Int. Con. on AI and Comp. Vis. (AICV2020)}, vol. 115. DOI: 10.1007/978-3-030-44289-7\texttt{\_}28

\bibitem{Kofinas18}
P. Kofinas, A.I. Dounis, and G.A. Vouros, "Fuzzy Q-learning for multi-agent decentralized energy management in microgrids," \emph{Applied Energy}, vol. 2019, pp. 53-67, Mar. 2018. DOI: 10.1016/J.APPENERGY.2018.03.017

\bibitem{Hang20}
P. Hang, C. Lv, Y. Xing, C. Huang, and Z. Hu, "Human-like decision making for autonomous driving: A noncooperative game theoretic approach", vol. 22, no. 4, pp. 2076-2087, Nov. 2020. DOI: 10.1109/TITS.2020.3036984

\bibitem{Yu2018}
H. Yu, H.E. Tseng, and R. Langari, "A human-like game theory-based controller for automatic lane changing," \emph{Transp. Res. Part C: Emerg. Tech.}, vol. 88, pp. 140–158, Feb. 2018. DOI: 10.1016/J.TRC.2018.01.016

\bibitem{An21}
Q. An , S. Cheng, C. Li, L. Li, and H. Peng, "Game theory-based control strategy for trajectory following of four-wheel independently actuated autonomous vehicles," \emph{IEEE Trans. on Int. Transp. Sys.}, vol. 70, no. 3, pp. 2196–2208, Mar. 2021. DOI: 10.1109/TVT.2021.3057161

\bibitem{Wang21}
M. Wang, Z. Wang, J. Talbot, J.C. Gerdes, and M. Schwager, "Game-theoretic planning for self-driving cars in multivehicle competitive scenarios," \emph{IEEE Trans. on Rob.}, vol. 37, no. 4, pp. 1313–1325, Jan. 2021. DOI: 10.1109/TRO.2020.3047521	
	
\bibitem{Jagodnik17}
K.M. Jagodnik, P.S. Thomas, A.J. van den Bogert, M.S. Branicky, and R.F. Kirsch, "Training an actor-critic reinforcement learning controller for arm movement using human-generated rewards," \emph{IEEE Trans. on Neural Sys. and Rehab. Eng.}, vol. 25, no. 10, pp. 1892-1905, Oct. 2017. DOI: 10.1109/TNSRE.2017.2700395

\bibitem{Baldominos19}
A. Baldominos, Y. Saezand, and P. Isasi, "Hybridizing evolutionary computation and deep neural networks: An approach to handwriting recognition using committees and transfer learning," \emph{Complexity}, vol. 2019, Mar. 2019. DOI: 10.1155/2019/2952304

\bibitem{Koller19}
T. Koller, F. Berkenkamp, M. Turchetta, J. Boedecker, and A. Krause, "Learning-based model predictive control for safe exploration and reinforcement learning," in \emph{Proc. of the 2018 IEEE Conf. on Decision and Cont. (CDC)}, Miami, USA, 2019. DOI: 10.1109/CDC.2018.8619572

\bibitem{Han20}
T. Han, S. Nageshrao, D.P. Filev, and U. Ozguner, "An online evolving framework for advancing reinforcement-learning based automated vehicle control," \emph{IFAC-PapersOnLine}, vol. 53, no. 2, pp. 8118–8123, July 2020. DOI: 10.1016/J.IFACOL.2020.12.2283

\bibitem{Hoel20}
C.-J. Hoel, K. Driggs-Campbell, K. Wolff, L. Laine, and M.J. Kochenderfer, "Combining Planning and Deep Reinforcement Learning in Tactical Decision Making for Autonomous Driving," \emph{IEEE Trans. on Intel. Veh.}, vol. 5, no. 2, pp. 294-305, June 2020. DOI: 10.1109/TIV.2019.2955905
	
\bibitem{Kiran2021}
B.R. Kiran, I. Sobh, V. Talpaert, P. Mannion, A.A. Al Sallab, S. Yogamani, and P. Pérez, "Deep reinforcement learning for autonomous driving: A survey," \emph{IEEE Trans. on Intel. Transp. Sys.} (Early Access), pp. 1-18, Feb. 2021. DOI: 10.1109/TITS.2021.3054625

\bibitem{Chen19}
J. Chen, B. Yuan, and M. Tomizuka, "Model-free deep reinforcement learning for urban autonomous driving," in \emph{Proc. of the 2019 IEEE Intel. Transp. Sys. Con. (ITSC)}, Auckland, NZ, 2019, pp. 2765-2771. DOI: 10.1109/ITSC.2019.8917306

\bibitem{Kendall19}
A. Kendall \textit{et al.}, "Learning to drive in a day," in \emph{Proc. of the 2019 Int. Con. on Rob. and Autom. (ICRA)}, Montreal, Canada, 2019, pp. 8248-8254. DOI: 10.1109/ICRA.2019.8793742

\bibitem{Isele19}
D. Isele, A. Nakhaei, and K. Fujimura, "Safe reinforcement learning on autonomous vehicles," in \emph{Proc. of the 2018 IEEE/RSJ Int. Con. on Intel. Rob. and Sys. (IROS)}, Madrid, Spain, 2019, pp. 6162-6167. DOI: 10.1109/IROS.2018.8593420

\bibitem{Bouton19}
M. Bouton, A. Nakhaei, K. Fujimura, and M.J. Kochenderfer, "Safe reinforcement learning with scene decomposition for navigating complex urban environments," in \emph{Proc. of the 2019 IEEE Intel. Veh. Symp. (IV)}, Paris, France, 2019, pp. 1469-1476. DOI: 10.1109/IVS.2019.8813803

\bibitem{Baheri20}
A. Baheri, "Safe reinforcement learning with mixture density network: A case study in autonomous highway driving," \emph{Results in Cont. and Optim.}, vol. 6, pp. 294-305, Mar. 2022. DOI: 10.1016/J.RICO.2022.100095

\bibitem{Wen20}
L. Wen, J. Duan, S.E. Li, S. Xu, and H. Peng, "Safe reinforcement learning for autonomous vehicles through parallel constrained policy optimization," in \emph{Proc. of the 2020 23\textsuperscript{rd} Int. Con. on Intel. Transp. Sys. (ITSC)}, Rhodes, Greece, 2020. DOI: 10.1109/ITSC45102.2020.9294262

\bibitem{Mirchevska18}
B. Mirchevska, C. Pek, M. Werling, M. Althoff, and J. Boedecker, "High-level decision making for safe and reasonable autonomous lane changing using reinforcement learning," in \emph{Proc. of the 2018 21\textsuperscript{st} Int. Con. on Intel. Transp. Sys. (ITSC)}, Maui, USA, 2018, pp. 2156-2162. DOI: 10.1109/ITSC.2018.8569448

\bibitem{Liu2021}
H. Liu, Z. Huang, and C. Lv, "Improved deep reinforcement learning with expert demonstrations for urban autonomous driving," 2021, [Online]. Available: https://arxiv.org/abs/2102.09243

\bibitem{Ferdowsi18}
A. Ferdowsi, U. Challita, W. Saad, and N.B. Mandayam, "Robust deep reinforcement learning for security and safety in autonomous vehicle systems," in \emph{Proc. of the 2018 21\textsuperscript{st} Int. Con. on Intel. Transp. Sys. (ITSC)}, Maui, USA, 2018, pp. 307-312. DOI: 10.1109/ITSC.2018.8569635

\bibitem{Xiong16}
X. Xiong, J. Wang, F. Zhang, and K. Li, "Combining deep reinforcement learning and safety based control for autonomous driving," 2016, [Online]. Available: https://arxiv.org/abs/1612.00147

\bibitem{Chen20}
D. Chen, L. Jiang, Y. Wang, and Z. Li, "Autonomous driving using safe reinforcement learning by incorporating a regret-based human lane-changing decision model," in \emph{Proc. of the 2020 American Cont. Con. (ACC)}, Denver, USA, 2020, pp. 4355-4361. DOI: 10.23919/ACC45564.2020.9147626

\bibitem{Shalev-Shwartz16}
S. Shalev-Shwartz, S. Shammah, and A. Shashua, "Safe, multi-agent, reinforcement learning for autonomous driving," 2016, [Online]. Available: https://arxiv.org/abs/1610.03295

\bibitem{Zhu20}
M. Zhu, Y. Wang, Z. Pu, J. Hu, X. Wang, and R. Ke, "Safe, efficient, and comfortable velocity control based on reinforcement learning for autonomous driving," \emph{Trans. Transp. Research Part C: Emerg. Tech.}, vol. 117, pp. 1-14, Aug. 2020. DOI: 10.1016/J.TRC.2020.1026625

\bibitem{Xu20}
X. Xu, L. Zuo, X. Li, L. Qian, J. Ren, and Z. Sun, "A reinforcement learning approach to autonomous decision making of intelligent vehicles on highways," \emph{IEEE Trans. on Sys., Man, and Cyber.: Sys.}, vol. 50, no. 10, pp. 3884-3897, Oct. 2020. DOI: 10.1109/TSMC.2018.2870983

\bibitem{Nayak17}
S. Nayak and M. Narvekar, "Real-time vehicle navigation using modified A* algorithm," in \emph{Proc. of the 2017 Int. Con. on Emerging Trends \& Innov. in ICT (ICEI)}, Pune, India, 2017, pp. 116-122. DOI: 10.1109/ETIICT.2017.7977021

\bibitem{ECE20}
ECE - United Nations, "Addendum 78: Regulation No. 79-03: Uniform provisions concerning the approval of: Vehicles with regard to steering equipment," \emph{InterRegs Ltd}, 2020, [Online]. Available: https://www.interregs.com/catalogue/details/ece-7903/regulation-no-79-03/steering-equipment/
	
\bibitem{Schulman17}
J. Schulman, F. Wolski, P. Dhariwal, A. Radford, and O. Klimov, "Proximal policy optimization algorithms," 2017, [Online]. Available: https://arxiv.org/abs/1707.06347

\bibitem{Dosovitskiy17}
A. Dosovitskiy, G. Ros, F. Codevilla, A. Lopez, and V. Koltun, "CARLA: An open urban driving simulator," in \emph{Proc. of the 1\textsuperscript{st} Annual Conf. on Rob. Learn. (PMLR)}, vol. 78, 2017, pp. 1-16.

\bibitem{Hill18}
A. Hill \textit{et al.}, "Stable Baselines," \emph{GitHub repository}, 2018, [Online]. Available: https://github.com/hill-a/stable-baselines
	
\bibitem{Andrychowicz20}
M. Andrychowicz \textit{et al.}, "What matters in on-policy reinforcement learning? A large-scale empirical study," in \emph{Proc. of the 9\textsuperscript{th} Conf. on Learn. Repres. (ICLR)}, Vienna, Austria, 2021.
	
\end{thebibliography}
\end{document}